# PR-NET: Leveraging Pathway Refined Network Structures for Prostate Cancer Patient Condition Prediction

Rui Li[1,2], Jia Liu[3], Xiaolong Deng[3], Xin Liu[3], Juncheng Guo[3], Wenyuan Wu[1*], Li Yang[1,3*], Lizhong Dai[3*]

[1]Chongqing Institute of Green and Intelligent Technology, Chinese Academy of Sciences, Chongqing 400714, China
[2] Chongqing School, University of Chinese Academy of Sciences, Chongqing 400714, China
[3] Sansure Biotech Incorporation, Changsha, Hunan 410000, China
[*]Corresponding authors：Wenyuan Wu, Chongqing Key Laboratory of Secure Computing for Biology, CIGIT, CAS, China. E-mail: wuwenyuan@cigit.ac.cn; Li Yang, Sansure Biotech Incorporation. E-mail: li.yang@sansure.com.cn; Lizhong Dai, Sansure Biotech Incorporation. E-mail: lizhongd@sansure.com.cn

## Abstract

The diagnosis and monitoring of Castrate Resistant Prostate Cancer (CRPC) are crucial for cancer patients, but the current models (such as P-NET) have limitations in terms of parameter count, generalization, and cost. To address the issue, we develop a more accurate and efficient Prostate Cancer patient condition prediction model, named PR-NET. By compressing and optimizing the network structure of P-NET, the model complexity is

---

 **Rui Li** is currently a PhD candidate at Chongqing Institute of Green and Intelligent Technology, Chinese Academy of Sciences, China. His research interests include privacy-preserving computation and machine learning.
**Jia Liu,** PHD, Senior Engineer, His main research interests include genomics and bioinformatics.
**Xiaolong Deng**, Master, bioinformatics Engineer, His main research interests include genomics and bioinformatics.
**Xin Liu,** bioinformatics Engineer, His main research interests include machine learning and bioinformatics.
**Juncheng Guo,** Master, bioinformatics Engineer, His main research interests include genomics and bioinformatics.
**Wenyuan Wu,** born in 1976, PhD, professor. His main research interests include artificial intelligence，bioinformatics methods and privacy-preserving computation.
**Li Yang**, born in 1984，PhD，professor. Her main research interests include nucleic acid test, next-generation sequencing and bioinformatics.
**Lizhong Dai,** PhD, professorate senior engineer. His main research interests include Molecular Biology, genomics and bioinformatics large-scale data analysis.

reduced while maintaining high accuracy and interpretability. The PR-NET demonstrated superior performance in predicting prostate cancer patient outcomes, outshining P-NET and six other traditional models with a significant margin. In our rigorous evaluation, PR-NET not only achieved impressive average AUC and Recall scores of 0.94 and 0.83, respectively, on known data but also maintained robust generalizability on five unknown datasets with a higher average AUC of 0.73 and Recall of 0.72, compared to P-NET's 0.68 and 0.5. PR-NET's efficiency was evidenced by its shorter average training and inference times, and its gene-level analysis revealed 46 key genes, demonstrating its enhanced predictive power and efficiency in identifying critical biomarkers for prostate cancer. Future research can further expand its application domains and optimize the model's performance and reliability.

## 1. Introduction

In recent years, cancer has emerged as the primary factor affecting human health. Taking prostate cancer as an example, it is one of the most common types of cancer among men , with a mortality rate of about 18.8% in the recent five years (Gundem, Van Loo et al. 2015), ranking as the fifth deadliest cancer worldwide. Depending on the location of the tumour and its stage of development, prostate cancer can be broadly categorized based on Primary Prostate Cancer (PPC) and Metastatic Prostate Cancer

(MPC). Patients with PPC generally have a positive prognosis (Zhao, Chen et al. 2020), with a five-year relative survival rate of over 99%. In contrast, patients with MPC face a poorer prognosis, particularly when the disease progresses to the castration-resistant phase, at which point mortality rates are often high (Armstrong, Lin et al., 2020). Therefore, monitoring and assessing the progression of Castrate Resistant Prostate Cancer (CRPC) in patients is critically important.

Interdisciplinary studies in artificial intelligence and medicine have gradually become research hotspots, particularly within the domain of oncology (Kourou, Exarchos et al. 2015; Esteva, Feng et al. 2022; Koh, Papanikolaou et al. 2022). There have been significant advancements in employing artificial intelligence for cancer research. In recent years, numerous studies have investigated the application of AI techniques, such as deep learning, to discern cancer dependencies, categorize cancer subtypes, and forecast cancer progression trends. Chih-Hsu Lin and colleagues developed a model using interpretable deep learning methods to determine cancer dependencies (Lin, C. H., & Lichtarge, O., 2021). Through the analysis of extensive cancer datasets, they uncovered critical factors and relationships pertaining to cancer progression. Runpu Chen and his team utilized high-dimensional genomic data alongside deep learning approaches to differentiate among various cancer subtypes (Chen, Runpu et al., 2020). Zhiqin Wang and associates employed a combination of genomic data and pathological imagery, applying a deep bilinear network to prognosticate breast cancer outcomes (Wang, Zhiqin et al., 2021).

However, previous research has mainly focused on unidimensional modeling and analysis of cancer. For organisms, a variety of metabolic pathways constitute a complex hierarchical network. Therefore, Elmarakeby et al (2021) have constructed a biological information deep learning model P-NET, based on sparse deep learning framework, biological information encoding, and interpretable algorithms. Compared to established models, it is expected to achieve better predictive performance. They trained and tested P-NET using a public dataset containing 1,013 samples and gene loci. P-NET has the advantages of high accuracy and interpretability. Through the innovative use of P-NET, relationships between genes such as *MDM4* and *FGFR1* and CRPC have been discovered, accelerating research on CRPC and driving progress in the field of modern oncology.

Nevertheless, P-NET does have some notable limitations. Firstly, its practicality is questionable The input data of P-NET is a set of gene loci information with a length of 27687, which requires the use of Whole Genome Sequencing (WGS) for DNA sampling of patients, and the price of individual WGS is around $8,000(BGI 2023), which is prohibitively expensive. Secondly, the model's generalization capability is less than ideal. P-NET's network architecture is exceedingly intricate, featuring a seven-layer network and a separate output layer for each layer. Only in the input layer, P-NET sets up 27,687 nodes with more than 70,000 model parameters. The complex structure will reduce the generalization ability of P-NET. Experiments show that the Recall value of P-NET is only about 0.5 when predicting patients with unknown types of prostate cancer, and the accuracy is even lower than 0.4, which is quite impractical in clinic.

In response to the aforementioned shortcomings of P-NET, we have undertaken a streamlined optimization of its network architecture. This process involved pruning superfluous nodes, restructuring the model, and reducing the number of parameters, which collectively enhanced the model's generalization capabilities and practical utility. Concurrently, we managed to decrease both the model's training and prediction times. Furthermore, we evaluated the refined P-NET across various datasets. The outcomes demonstrate an improvement in the recall rate of the optimized P-NET.

Overall, we have developed a predictive model for prostate cancer patient outcomes, termed PR-NET (Pathway-Refined NET), which outperforms P-NET in terms of accuracy and efficiency. In comparison to P-NET, the operational cost of PR-NET has been reduced by roughly 90%, although this reduction may vary with different sequencing products. The training duration for the model has lengthened by a factor of 2.67, a figure contingent upon the computational equipment utilized during this study's experiments. In terms of predicting unknown datasets, PR-NET's recall rate can exceed 0.7, marking an approximate 33% enhancement.

## 2. Materials and Methods

### 2.1 Data Acquisition and Preprocessing

Our datasets consists of five publicly available prostate cancer datasets: Prad-p1000, Prad-msk-stopsack-2021, Prad-pik3r1-msk-2021, Prad-cdk12-mskcc-2020, and Prad-mskcc-2017. Detailed statistics of the datasets are summarized in Table 2.1.

- Data1 Prad-p1000(Armenia, Wankowicz et al. 2018)
- Data2 Prad-msk-stopsack-2021(Stopsack, Nandakumar et al. 2022)
- Data3 Prad-pik3r1-msk-2021(Chakraborty, Nandakumar et al. 2022)
- Data4 Prad-cdk12-mskcc-2020(Nguyen, Mota et al. 2020)
- Data5 Prad-mskcc-2017(Hulsen 2019)

**Table 2.1** Statistical results of the data set

| ID | Type | CRPCs | Primary Cancers | Sequencing Technology Platform |
|---|---|---|---|---|
| Prad-p1000 | Prostate Cancer | 333 | 680 | WES |
| Prad-msk-stopsack-2021 | Prostate Cancer | 990 | 1062 | MSK-IMPACT panel |
| Prad-pik3r1-msk-2021 | Prostate Cancer | 592 | 825 | MSK-IMPACT panel |
| Prad-cdk12-mskcc-2020 | Prostate Cancer | 608 | 857 | MSK-IMPACT panel |
| Prad-mskcc-2017 | Prostate Cancer | 228 | 276 | MSK-IMPACT panel |

For the five datasets listed in Table 2.1, our analysis concentrates on gene mutations, copy number amplification, copy number deletion, and patient response. In order to handle missing values, we have imputed them with zeros. Moreover, we have defined CPRC as positive samples and primary type as negative samples. This allows us to divide the datasets into training and testing sets. In order to further evaluate the generalization capability of each model, we preprocessed all datasets to ensure consistent dimensions. For a comprehensive account of the data preprocessing protocol, please consult the code implementation provided in the Supplementary Information.

**2.2 Constructing the PR-NET Model**

One of the key challenges in building the P-NET model lies in dealing with its high complexity, mainly due to the large number of model parameters. Notably, the input layer alone accounts for approximately 94% of the total parameters. Hence, our primary focus is on optimizing the input layer.

Initially, we train the P-NET using the prad_p1000 dataset. Upon completion of training, we apply DeepLIFT (Shrikumar, A., Greenside, P., et al. 2017) to determine the importance scores for all genetic loci in the input layer. We then rank the genetic loci in descending order by their importance scores.

Next, we refine the model by discarding all gene loci that have an importance score of zero, thereby keeping only the significant loci. Continuing with this process, we begin with the top six gene loci that exhibit the highest importance scores and progressively include additional loci. The P-NET is retrained throughout this process, incorporating each new set of loci. We identify a particular set of genetic loci, comprising loci 37 (0.4%), 45 (0.487%), 46 (0.5%), 47 (0.502%), 56 (0.6%), 89 (1%), and 3,751 (40%). These specific loci are chosen based on their proportional significance within the dataset. Following this selection, we extract the relevant data from the prad_p1000 dataset for further retraining of the P-NET.

Ultimately, we meticulously observe the model's performance to pinpoint the peak of its capability. At this optimal point, we define the set of genetic loci as the most favorable combination and henceforth designate the resulting model as PR-NET.

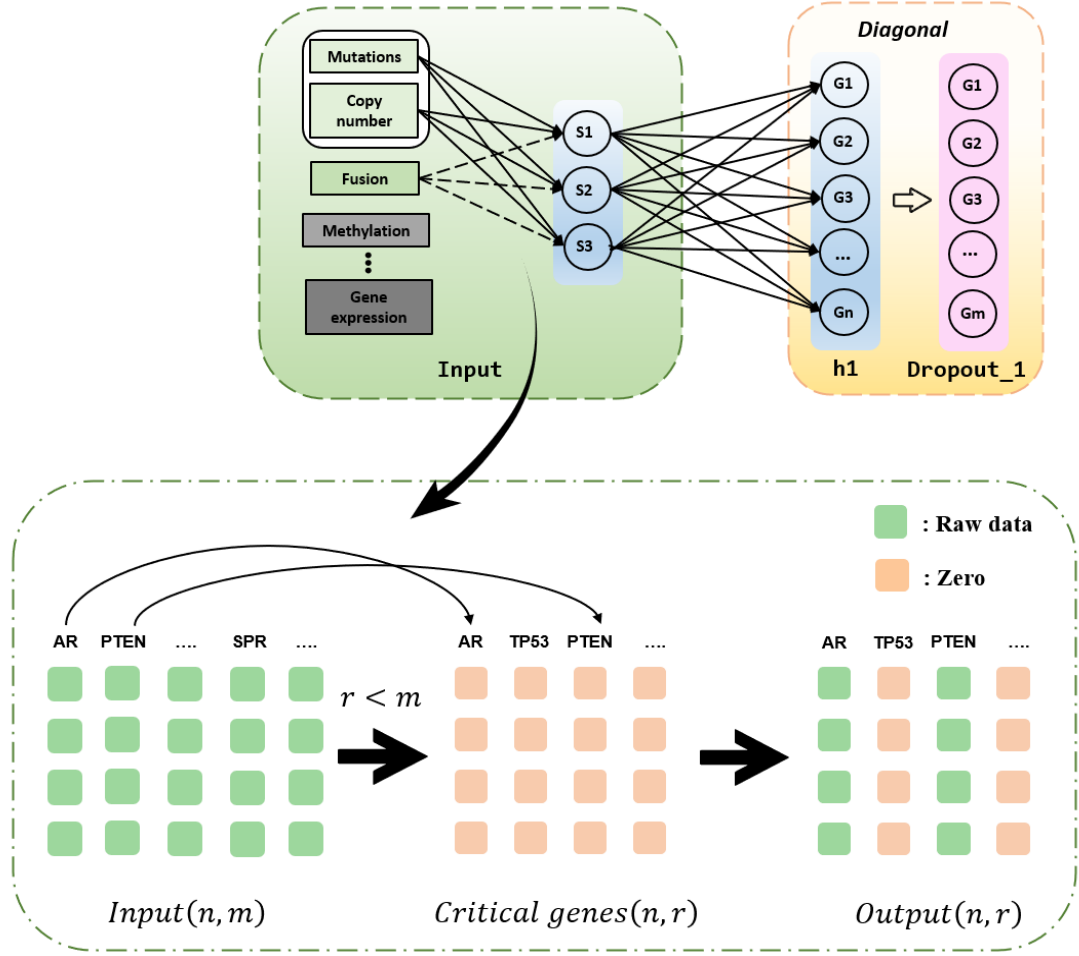

Figure 2.1 illustrates the structure of the input layer for the PR-NET model.

(For a dataset $data_0$ consisting of $n$ samples and containing $m$ genetic loci $(m > r)$, we extract the data corresponding to the optimal combination of genetic loci before it enters the input layer. Consequently, a refined dataset, referred to as $data_1$, is manifested, featuring $n$ samples and $r$ genetic loci. This refined dataset, $data_1$, assumes the role of the actual input data for subsequent training and prediction tasks.)

It is imperative to acknowledge that during the prediction process of PR-NET, we introduce two sets to punctuate the genetic loci information. $G_0$ represents the comprehensive collection of all genetic loci information within $data_0$, while $G_1$ denotes the set of optimal genetic loci. The proposed PR-NET framework incorporates a set of soft constraints on the genetic loci present in $data_0$, which are as follows:

$$\begin{cases} |G_0| \geq |G_1| \\ G_1 \subseteq G_0 \end{cases} \tag{2.6}$$

The primary objective of the aforementioned constraint is to guarantee the inclusion of the maximum number of genetic loci in the dataset to be predicted.

**2.3 Rationality Analysis of the Optimization Method in PR-NET**

In section 2.2, we validate the optimization method employed for selecting the optimal genetic loci combination for the PR-NET model. This validation encompasses the analysis of importance score statistics, correlation tests, and mutual information analysis from various perspectives. Through these analyses, we ascertain the significance of the chosen loci, their correlation with patient response, and the efficacy of the optimization approach.

**2.4 Performance Testing of PR-NET**

Following a comprehensive analysis and validation of the optimization method used, we proceed with performance testing of PR-NET. Our goal is to conduct comparative tests across various models, including decision trees, regularized logistic regression, random forest classifiers, linear and non-linear kernel-based models, as well as P-NET and PR-NET themselves. The testing will be organized into three categories: a performance comparison based on the different models, a comparison of generalization abilities using the full set of genetic loci, and a comparison using only the optimal set of genetic loci. The data for all tests will come from the dataset referenced in section 2.1. Lastly, we assess the computation times for both P-NET and PR-NET.

## 3. Experimental Results

Through our experiments, we have demonstrated the rationality of the selected genetic loci. The 46 filtered genetic loci are used as the optimal combination, optimizing the input layer of P-NET and constructing PR-NET. In terms of model performance, PR-NET exhibits superior generalization ability compared to P-NET.

**3.1 Results of Optimal Genetic Loci Selection**

Figure 3.1 presents the results obtained from repeating the training of P-NET for 5 rounds using the optimization method outlined in section 2.2. It is evident from the figure that the utilization of redundant genetic loci can have an impact on model performance. In order to gain further insights, we conducted a statistical analysis and comparison of the number of genetic loci and model parameters in the input layer before and after optimization, as depicted in Figure 3.2. The results indicate that after optimization, there was an approximate reduction of 99% in the number of genetic loci and an approximate reduction of 85% in the number of model parameters. It is worth noting that this significant reduction in model parameters is expected to yield substantial reductions in both training and prediction time. These findings underscore the rationale behind the optimization method employed and the selection of genetic loci.

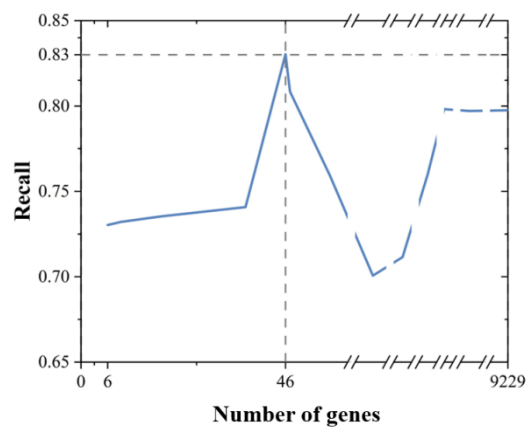

Figure 3.1: Variation Trend of Recall with Different Numbers of Genes Selected

（When selecting 46 genetic loci, the Recall reaches its peak..）

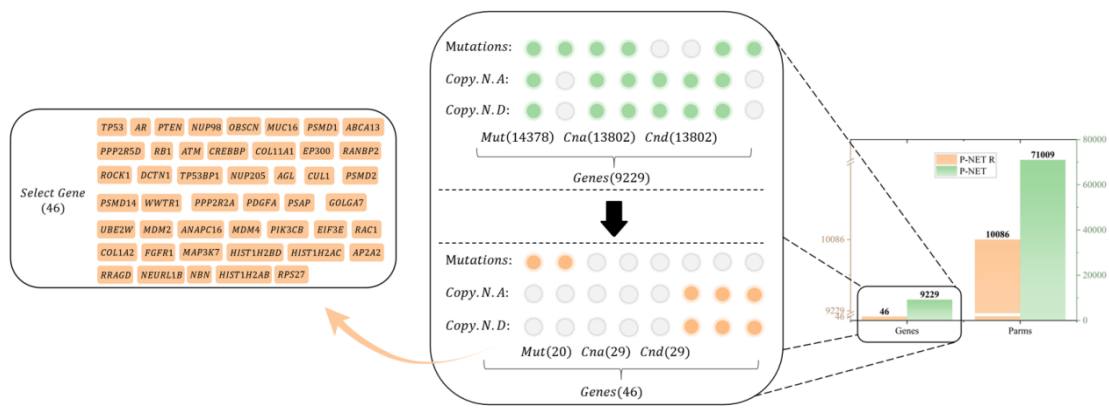

Figure 3.2: Specific Information of Key Gene Loci

（Number of gene loci optimized from 9,229 to 46）

## 3.2 The Rationality Analysis Results of the Optimization Method

The importance scores of gene loci were computed and analyzed utilizing the P-

NET trained with DeepLIFT. The distribution of these scores was subsequently examined across three intervals: [0,0], (0,1], and (1,∞). As depicted in Figure 3.3, the prevailing proportion of gene loci was deemed irrelevant throughout the training process of the P-NET model for the present problem. This outcome is in line with our initial expectations, thereby indicating that only a select few key gene loci contribute significantly to prostate cancer classification. Moreover, a comparison was made between the importance scores and rankings of the optimized gene loci before and after the optimization process, as illustrated in Figure 3.4. It is evident that the rankings of the top gene loci remained largely unaltered, while the importance scores exhibited substantial variations. Consistent with our expectations, P-NET demonstrated improved ability to identify "important features" through this optimization.

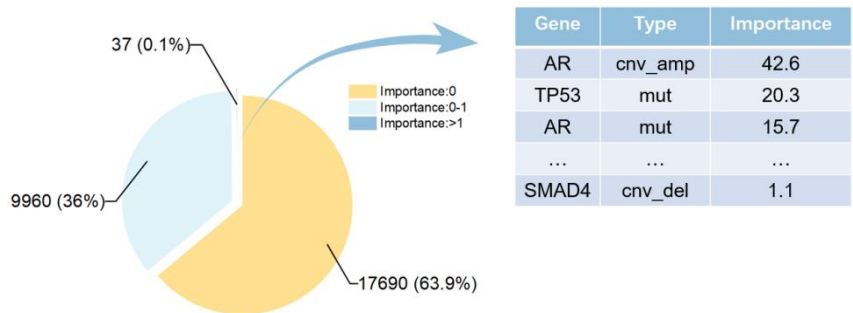

Figure 3.3: Statistical Results of Gene Importance Scores

（In the three categories of information across 9,229 gene loci (each gene locus includes Mutations, Copy Number Amplification, and Copy Number Deletion), over 60% of the gene loci are distributed within the interval [0,0]. Approximately 36% of the gene loci are distributed within the interval (0,1], while only 37 gene loci are distributed within the interval (1, +∞).）

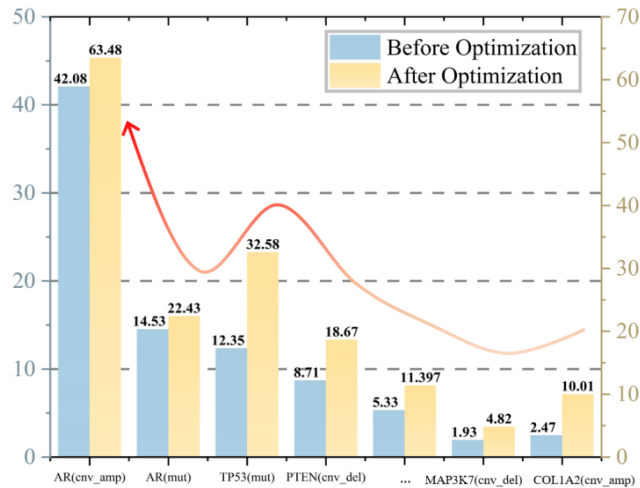

Figure 3.4: Comparison of Importance Scores and Rankings for Top Gene Loci before and after Optimization

(After optimization, the rankings of the top gene loci remained mostly unchanged, but there were significant variations in their importance scores. For instance: $AR(cnv\_amp)$: 42.08 → 63.48, $TP53(mut\_imp)$: 12.35 → 32.58）

Correlation Analysis: A sequential assessment was performed to analyze the correlation between the top gene loci and the variable "Response." The gene loci were then ranked according to their correlation strength, with the detailed results presented in Figure 3.5. Our research indicates that gene loci with higher importance scores generally show a stronger correlation with "Response." These observations affirm the validity of the optimization method applied in our study.

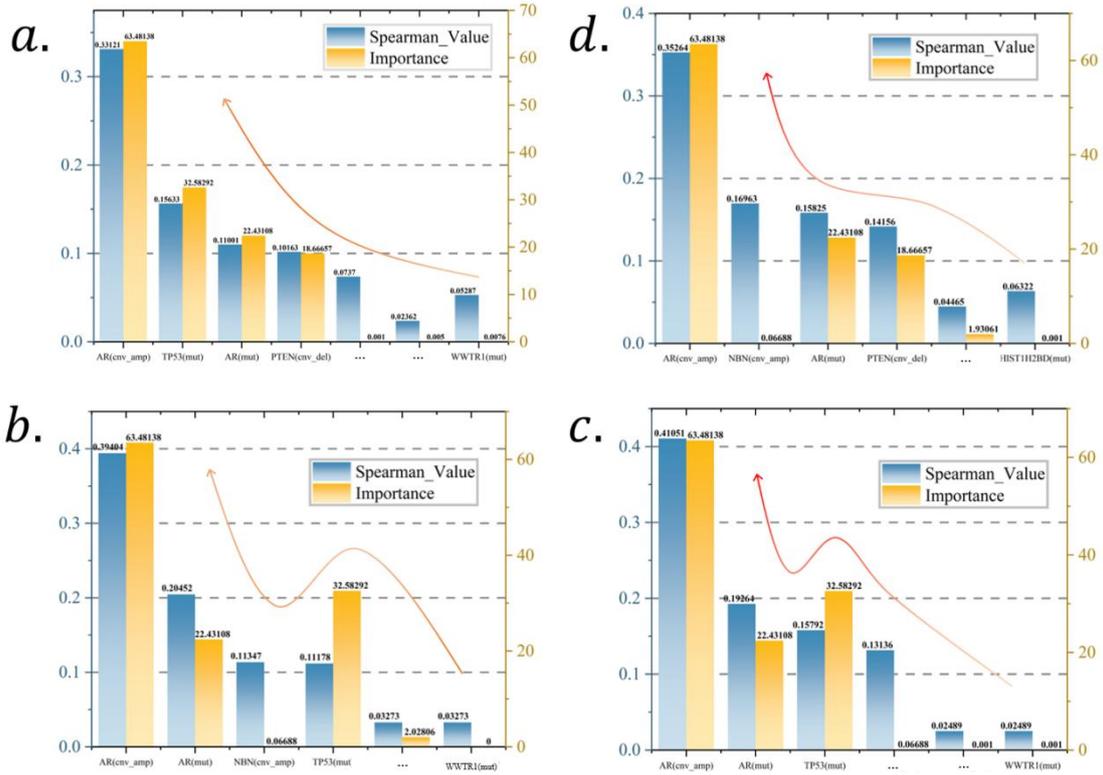

Figure 3.5: Correlation Analysis Results of Top Gene Loci with "Response"

(From Figures $a$ to $d$, we present the correlation analysis results of the top gene loci with "Response" in $Data_2$ to $Data_5$. Particularly, "$AR(cnv\_amp)$" exhibits the strongest correlation with "Response" across all four datasets, and it also has the highest importance score. However, there are also some gene loci (e.g., $b.NBN(cnv\_amp)$ $c.NBN(cnv\_amp)$) that show strong correlation with "Response" but have relatively lower importance scores. Additionally, there are certain gene loci (e.g., $b.TP53(mut)$ $c.TP53(mut)$) that have higher importance scores despite their weaker correlation with "Response." This could be attributed to the variations in the datasets used.)

Mutual Information Analysis: For $Data_i$ $(i = 2,3,4,5)$, we computed the mutual information with respect to "Response" for each gene locus and ranked them accordingly. The specific results are shown in Figure 3.6. We observed that gene loci with higher mutual information scores generally correspond to those with higher importance scores. This correlation provides additional confirmation of the soundness of the optimization method used in our research.

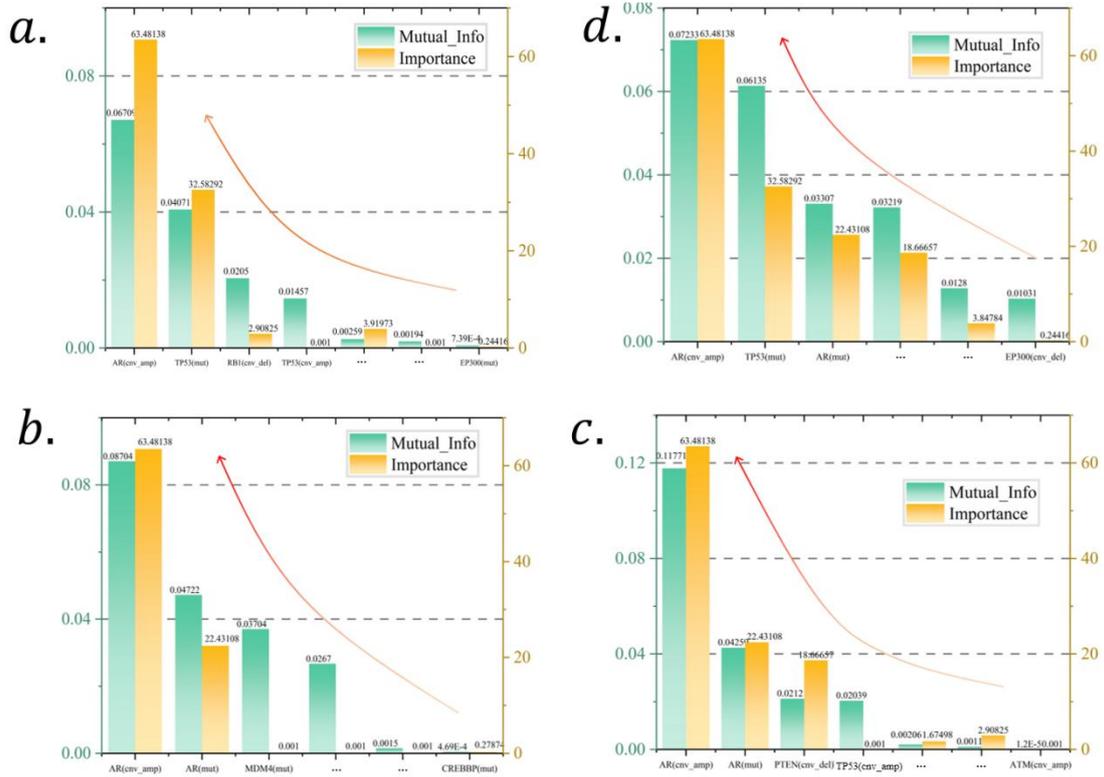

Figure 3.6: Mutual Information Calculation Results of Key Genes with "Response"

（From Figures $a$ to $d$, the mutual information results of the top gene loci with "Response" in $Data_2$ to $Data_5$ are displayed, respectively.）

## 3.3 Model Performance Testing

The performance comparison test based on $Data_1$ models involved training and testing of various models. The results, as depicted in Figure 3.7, indicate that PR-NET exhibits exceptional model performance and surpasses P-NET in terms of overall performance.

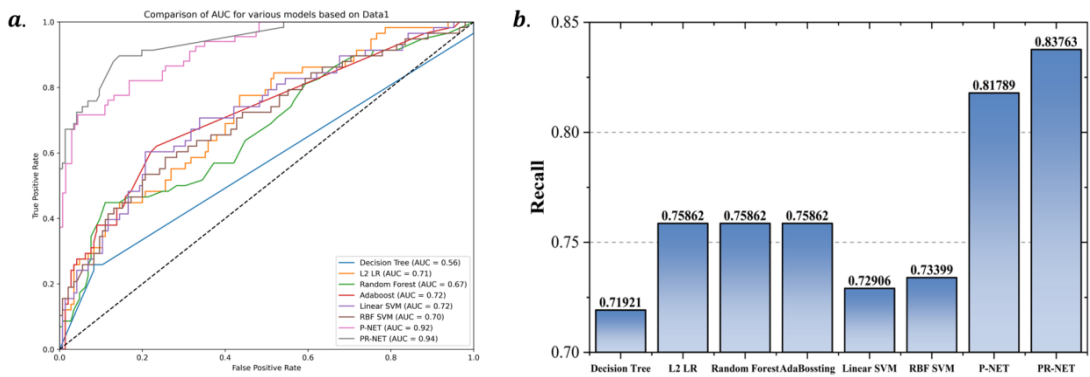

Figure 3.7: Comparative Analysis of Model Performance on $Data_1$

（During the evaluation conducted on $Data_1$, PR-NET showcased remarkable performance by attaining an AUC of 0.94 and a recall rate of 0.83, surpassing not only P-NET but also six other conventional machine learning models.）

Figure 3.8 presents the results of a comparative analysis evaluating the generalization capability based on whole-genome loci. In this study, we randomly selected 202, 404, 606, and 808 samples from $Data_1$ for training various models. Subsequently, we expanded $Data_i$ ($i = 2,3,4,5$) into $Data_j^E$ ($j = 2,3,4,5$), and predicted all samples in $Data_j^E$. Notably, PR-NET consistently maintained an accuracy of over 0.70, indicating its ability to ensure high precision when confronted with unknown data.

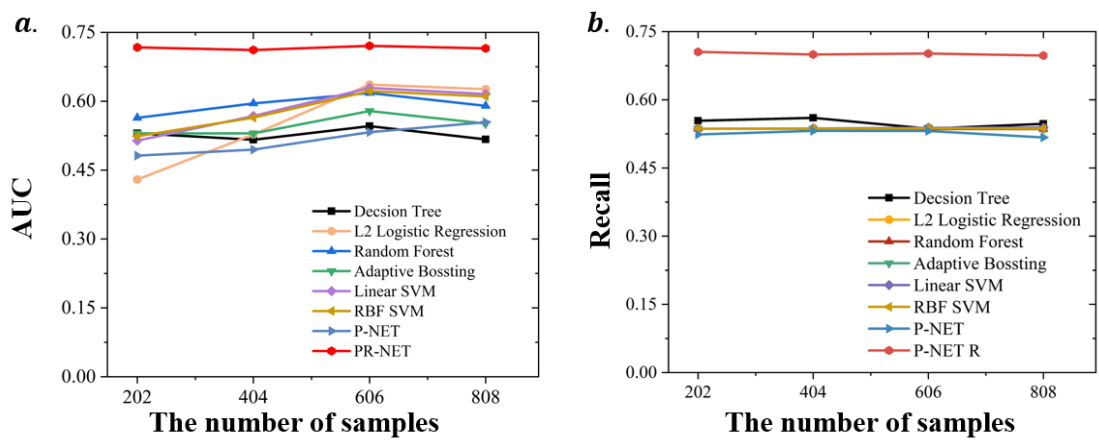

Figure 3.8: Comparative Analysis of Generalization Capability Based on Whole-Genome Loci

（PR-NET demonstrates an AUC and recall rate close to 0.75, surpassing both P-NET and other classical machine learning models.）

To validate the universality of our optimization method in improving model generalization capability, we first selected the optimal genomic loci from the five datasets, resulting in $Data_i^R$ ($i = 1,2\ldots,5$). Subsequently, all models were trained on $Data_1^R$ and further subjected to comparative testing on $Data_j^R$ ($j = 2,3,4,5$). The test results, as shown in Figure 3.9, demonstrate that PR-NET exhibits superior generalization capability, reaffirming its effectiveness in diverse scenarios.

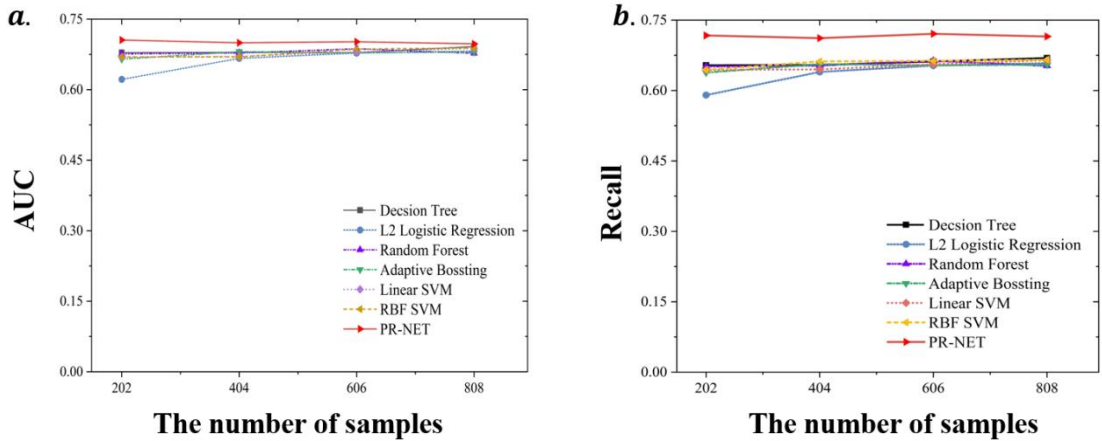

Figure 3.9: Comparative Analysis of Generalization Capability Based on Key Genomic Loci

(PR-NET (Ours) demonstrates higher AUC and recall rates in the generalization capability test compared to other machine learning models.)

On the same hardware level, based on Prad_P1000, we repeated the training and prediction of P-NET and PR-NET 5 times, using "average training time" and "average prediction time" as evaluation metrics. The experimental results are shown in Table 3.1.

Table 3.1: Comparison of computational time between PR-NET and P-NET

|  | Model | Average Time/s | Variance | Standard Deviation |
|---|---|---|---|---|
| Training time | P-NET | 520.009 | 576.958 | 24.019 |
|  | PR-NET | 195.499 | 66.926 | 8.180 |
| Inference time | P-NET | 0.534 | 0.002 | 0.053 |
|  | PR-NET | 0.146 | 7.90E-04 | 0.028 |

It is evident that PR-NET surpasses P-NET in both training and inference times. Specifically, PR-NET's average training time is approximately 1.6 times shorter than that of P-NET, and it demonstrates lower variance and standard deviation. Similarly, for inference time, PR-NET's average duration is about 2.6 times shorter, with reduced variance and standard deviation as well.

These experimental results suggest that PR-NET not only has a significant computational time advantage over P-NET but also offers greater stability.

In summary, the PR-NET model we have constructed not only outperforms P-NET in terms of model performance and practicality, but also exhibits significant superiority over six other classical machine learning models. This provides further evidence of the universal effectiveness of our optimization method in enhancing the generalization capability of machine learning/deep learning models.

## 4. Discussion

In this study, we conducted an optimization of the P-NET model based on previous literatures. Through exploring the importance ranking at the gene level, we effectively identified 46 gene loci, resulting in the highest recall for P-NET in both horizontal and vertical comparisons. Additionally, this reduction in the number of gene loci at the input layer of P-NET yielded a significant cost reduction of approximately 96.364% and a decrease in the number of model parameters by 85.796%. Overall, the refined model, called PR-NET, demonstrated superior performance compared to the original model across various evaluation metrics, while still preserving the significance of importance ranking.

Among the 46 gene loci selected during the optimization process, the top-ranked genes play pivotal roles in cancer. For instance, the *AR* gene encodes the androgen receptor, which exerts a crucial influence on hormone-related cancers, such as breast and prostate cancer. It actively promotes the growth and proliferation of tumor cells, rendering it a prime target for hormone therapy. The *TP53* gene codes for a tumor suppressor protein, which plays a vital role in safeguarding genomic integrity within cells. Mutations in the *TP53* gene demonstrate a close association with the development of several cancers(Wang, H., Guo, M., Wei, H. et al. 2023). The *PTEN* gene codes for a phosphatase that governs key processes, including cell proliferation, survival, and metabolism(Lee, YR., Chen, M. & Pandolfi, P.P. 2018.). By suppressing the *PI3K/AKT* signaling pathway, it effectively inhibits the growth and proliferation of tumor cells. Mutations in the *PTEN* gene are significantly linked to the occurrence and progression of various malignant tumors, like breast, prostate, and colorectal cancers.

The PR-NET outperforms the commonly used models across six datasets, demonstrating its effectiveness in cost reduction, efficiency enhancement, and improved learning capabilities on key features. Our method not only offers new insights and methods for gene locus selection and cancer diagnosis but also opens up new research opportunities and practical applications in related fields. Future studies can extend the applications to other types of cancers or diseases, further validating the versatility and robustness of the optimized model. Moreover, in today's information age, the protection of personal data has become an increasing concern. Homomorphic encryption can effectively solve the problem of privacy disclosure in deep learning. However, operations on encrypted data, especially comparison operations, are time-consuming and often can't be directly executed; instead, they require the support of secure protocols for implementation. Therefore, when integrating homomorphic encryption with deep learning models, efficiency becomes a critical factor. Fortunately, PR-NET achieves an 85% reduction in model parameters compared to P-NET, and its inference process does not involve comparison operations. This reduction significantly simplifies the complexity of the encrypted PR-NET model, enhances its computational efficiency, and improves the practicality of implementing PR-NET in an encrypted state.

Despite of optimized significant achievements of our optimization work, there are still some directions that can be explored. For instance, a large-scale dataset is still required to further ensure its reliability and scalability. Additionally, we will focus on enhancing the overall performance and robustness of our model. Furthermore, we are dedicated to actively exploring the application of our model in other clinical practices, and facilitating its translation and adoption to yield benefits in the field of cancer diagnosis and treatment.

## Author contributions

W.Y.W., L.Y. and J.L. conceived the study and supervised the project. R.L. implemented the model and performed analyses. X.L.D., X.L. and J.C.G. curated data sets ,evaluated the benchmarks and discussed the results and implications. All authors participated in the discussion of the manuscript.

## Supplementary data

Supplementary data are available at https://academic.oup.com/bib.

## Data and code availability

All code and data are available at https://github.com/LeeRay629/PR-NET2024

## Declaration of competing interest

The authors declare that they have no known competing financial interests or personal relationships that could have appeared to influence the work reported in this paper.

## Funding

This work was supported by the Major Research and Development Program of the Ministry of Science and Technology of China [2020YFA0712300] and Chongqing Natural Science Foundation [CSTB2023YSZX-JCX0008, 2022YSZX-JCX0011CSTB, cstc2021yszx-jcyjX0004].